# Gender bias in magazines oriented to men and women: a computational approach

Diego Kozlowski[1], Gabriela Lozano[2*], Carla M. Felcher[3*], Fernando Gonzalez[4], Edgar Altszyler[5,6]

[1] Faculty of Science, Technology and Medicine, University of Luxembourg, L-4364 Esch-sur-Alzette, Luxembourg. (ORCID: 0000-0002-5396-3471). E-mail: diego.kozlowski@uni.lu (corresponding author)

[2] Interdisciplinary Institute of Gender Studies- University of Buenos Aires.

[3] Instituto de Fisiología, Biología Molecular y Neurociencias (IFIBYNE), CONICET-Universidad de Buenos Aires, Buenos Aires, Argentina

[4] Maestría en Explotación de Datos y Descubrimiento del Conocimiento, FCEyN, Universidad de Buenos Aires..

[5] Departamento de Computación, FCEyN, Universidad de Buenos Aires.

[6] Instituto de Investigación en Ciencias de la Computación, CONICET-UBA, Argentina E-mail: ealtszyler@dc.uba.ar (corresponding author)

[*] These authors contributed equally

## Abstract

Cultural products are a source to acquire individual values and behaviours. Therefore, the differences in the content of the magazines aimed specifically at women or men are a means to create and reproduce gender stereotypes. In this study, we compare the content of a women-oriented magazine with that of a men-oriented one, both produced by the same editorial group, over a decade (2008-2018). With Topic Modelling techniques we identify the main themes discussed in the magazines and quantify how much the presence of these topics differs between magazines over time. Then, we performed a word-frequency analysis to validate this methodology and extend the analysis to other subjects that did not emerge automatically. Our results show that the frequency of appearance of the topics *Family*, *Business* and *Women as sex objects*, present an initial bias that tends to disappear over time. Conversely, in *Fashion* and *Science* topics, the initial differences between both magazines are maintained. Besides, we show that in 2012, the content associated with *horoscope* increased in the women-oriented magazine, generating a new gap that remained open over time. Also, we show a strong increase in the use of words associated with feminism since 2015 and specifically the word *abortion* in 2018. Overall, these computational tools allowed us to analyse more than 24,000 articles. Up to our knowledge, this is the first study to compare magazines in such a large dataset, a task that would have been prohibitive using manual content analysis methodologies.

*Conflicts of interest/Competing interests:*

The authors declare that they have no conflict of interest.

*Funding*

This work has been supported by the Doctoral Training Unit *Data-driven computational modelling and applications* (DRIVEN), which is funded by the Luxembourg National Research Fund under the PRIDE programme

(PRIDE17/12252781).

*Acknowledgments*

The authors want to thank Alfredo Rolla and Jorge Federico Cubells for their valuable help in this project.

# INTRODUCTION

Representations of gender in any society cannot be understood without considering the political and cultural intersections that comprise the context in which they were produced and maintained (Butler 1998). Gender is a historical construction (Scott 1986) deeply embedded in the use of discourse (Del-Teso-Craviotto 2006). Culture is a network of meanings that produce social realities (Serret, 2001; Lama, 2013). The relevance of popular media for the production and reproduction of stereotypes in societies has been widely studied (Foucault 1978). One of the most important theoretical frameworks on the role of the media as a stereotype reproductive agent is the Social Cognitive Theory (Bandura et al. 1963). This theory states that behavioural patterns and attitudes can be acquired by observing symbolic models, both those observed in real life and those shown in popular media. Therefore, portrayal of women and men in stereotypical roles and frames in mass media serves as a source of generation and reproduction of gender stereotypes (Del-Teso-Craviotto 2006).

Over the years, several authors have reported the reproduction of stereotyping associations by mass media such as magazines (Goffman 1979, Murillo et al, 2010), movies (Gálvez et al., 2019; Gilpatric 2010, Neuendorf et al. 2010), newspapers (Matud et al. 2011), radio (Eisend 2010), music (Turner 2011; Spataro 2013), and television (Eisend 2010; Das 2011; Desmond and Danilewicz 2010; Hether and Murphy 2010; Koivula, 1999). A revision on content analysis in mass media has been reported by Collins (2011) and Rudy (2010).

Another evidence of stereotyping association is the existence of differential content presented in the media expected to be consumed by men or women respectively. This bias strengthens and perpetuates differences between the symbolic role models of genders (Del-Teso-Craviotto 2006).

The use of coding schemes for content analysis represents a standard approach for studying stereotyped roles (Neuendorf 2011), and for several decades, have been used to quantify elements of static categories. However, vertiginous social changes may turn categories designed in the past outdated in short periods, therefore new categories may be required (Zotos & Tsichla 2014).

In the present study, we analyse the differences in content between women and men-oriented magazines. As source, we used *OHLALÁ* and *Revista Brando*, two Argentinean magazines from the same editorial board that are targeted at women and men, respectively. The fact that they come from the same editorial gives a unique opportunity for comparison. Here, we apply computational techniques to identify the content in 24,000 articles between 2008 and 2018 from these magazines, and then, we analyse content evolution and differences between them over the years.

## METHODS

In this section we describe the technical approaches of content analysis used to conduct this research. First, we present the dataset compiled and its characteristics. Second, we describe a Topic Modelling technique employed for automatic detection of topics in the dataset. Third, we outline a Word Frequency Analysis used to validate the automatic content detection. Finally, we describe the dataset preparation for these technical approaches. Given that the articles are originally in Spanish, the content analysis was performed in this language. A translation to English of the selected words and topics is provided.

### 1. Data collection

The corpus compiled for this paper consist of 6.060 articles downloaded from *Revista Brando*, and 18.082 articles from *OHLALÁ* magazine, published between 2008 and 2018 (Magazine links, accessed in 2018). Figure 1 shows the distribution of articles over time for each magazine. All the available articles online were gathered. According to the commercial information of each magazine (Comercial LA NACION, accessed in 2020) *Revista Brando*'s target are men between 30 and 50 years and with high purchasing power. It has a net sale of 10,000 copies, and 4,500 subscribers. *OHLALÁ*'s target are women between 25 and 45 years and also with high purchasing power. It has a net sale of 37,500 copies and 15,000 subscribers.

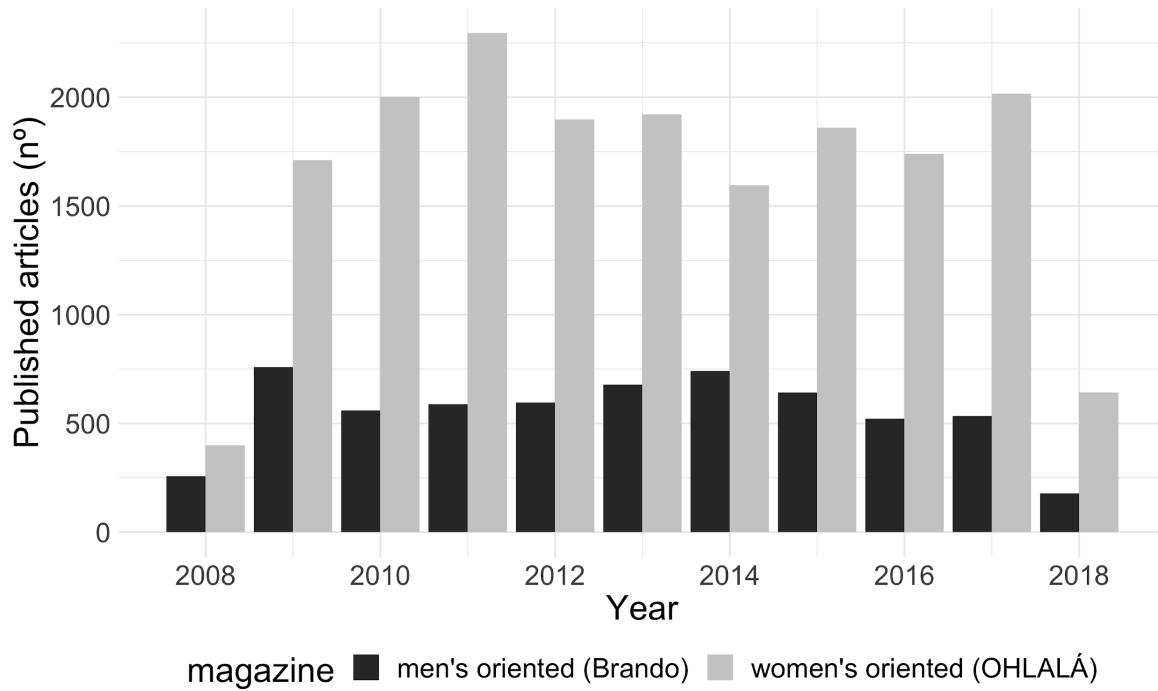

**Fig. 1**: Articles published by year for men and women-oriented magazines, Brando and OHLALÁ.

## 2. Topic Modelling

We implemented a Topic Modelling technique, the Latent Dirichlet Allocation (LDA) (Blei et al. 2003). The main function of LDA is to automatically identify the topics that best describe a dataset. This approach provides the opportunity of an unbiased analysis, as the topics arise from the corpus itself and not from the choice of the researcher. Since this model has been extensively explained by Blei et al. (2003), it will only be described briefly herein. Given a corpus of documents, LDA makes the assumption that there is a set of fixed topics that describe the content. LDA models each article as a distribution over topics, and each topic as a distribution over words. This can be thought of as the process of writing an article, in which the author first chooses the main themes, and then that themes condition which words are more likely to appear in the manuscript. After processing the texts, LDA automatically computes the probability distribution for each topic on each article. Then, each topic obtained by the LDA can be represented by a list of its most probable words. Given that list, a researcher in the field needs to analyse the lists of words and assign a label to each emergent topic. For example, a list composed of ten words including "children", "mother", "mom" and "father" among others, can be manually labelled as "traditional family".

In the present study, we fed the LDA with all the articles from both magazines and instructed the model to construct a high number of topics, 100. Given the table containing the 100 emerging topics and the 10 most probable words defining each of them, we selected and tagged the specific topics that we considered of interest regarding gender stereotypes, without prior knowledge of how they were distributed across magazines (Supplementary Table 1).

Then, for each magazine, we calculated the proportion in which each labelled topic is present for each year ($P_{M,Y} (T_i)$). As an example, this would be read as the proportion of the "family" topic in OHLALÁ magazine in 2010.

This is accounted by

$$P_{M,Y} (T_i) = \frac{\sum_d P_{d,M,Y}(T_i)}{\#d_{M,Y}} \qquad (1),$$

Where $P_{d,M,Y} (T_i)$ is the probability of the topic $i$ ($T_i$) for document $d$ in magazine $M$ and year $Y$, and $\#d_{M,Y}$ is the number of documents for magazine $M$ and year $Y$.

The main and novel contribution of Topic Modelling is the large-scale automatic identification of recurring topics in texts. In addition, it allows quantifying the presence of these topics among the magazines, thus revealing the differences present in magazines aimed at women and men.

It should be mentioned that not all inferred topics are interpretable (Chang et al.,

2009), therefore an expert from the field is required to select the topics that she or he considers most coherent and most relevant for the hypothesis testing. Another relevant factor to consider is that inferred topics can contain unexpected associations that arise from the discursive context of the corpus (Mohr and Bogdanov, 2013). To validate the LDA analysis in this study, we also performed a Words Frequency analysis as described below.

### 3. Words Frequency Analysis

We used the standard approach based on word counting. First, a list of unambiguously, manually selected, words was assembled to represent each topic. Second, the frequency of occurrence of these words for each year and each magazine was calculated. The frequency $F$ of occurrences of the word-list $l$, in the year $Y$ and magazine $M$ was calculated as

$$F_{M,Y}(l) = \frac{\#W_{l,M,Y}}{\#W_{*,M,Y}} \qquad (2),$$

Where $\#W_{l,M,Y}$ is the number of occurrences of words in the wordlist $l$ in magazine M and year Y, while $\#W_{*,M,Y}$ is the total number of words in magazine M for that given year (Michel et al. 2011).

Finally, for each word list, we compared the time series of the $F_{M,Y}(l)$ frequencies of both journals, and used a fisher-exact test to assess if differences are significant between both magazines. The words that have polysemy were excluded from the lists constructed to avoid overlapping in word meanings. Given that the vocabulary used in the magazines is composed of thousands of words, and that we are building lists of words of undefined sizes, the possible combinations of words is enormous. Therefore there are several thousands of potential tests. This generates a statistical problem known as "multiple comparison problem" meaning that it is possible to find a combination of words that provide statistically significant differences where there are none, just by chance. This has to be avoided, otherwise it could become a source of misleading conclusions. Thus, the lists have to be carefully constructed and theoretically justified before experimentation. In this work, we construct word lists based on the Topic Modelling results, thus avoiding this problem. We also build three lists of words that didn't appear automatically in the Topic Modelling results, but we consider they pose relevant theoretical issues: the rise of the feminist movement in Argentina during the period of study, together with the discussion in the Argentinean congress of the legalization of abortion, and finally the increasing consumption of pseudoscience, represented in the zodiacal signs.

**Dataset Preparation**

Each article was labelled with its release year *t*, and the magazine of origin (OHLALÁ or Brando). The stop words, like *the, they, to, etc,* were removed using the Spanish stop words list in the NLTK Python package (Bird et al., 2009). The list of the 500 most frequent words in Spanish was reviewed and a selection of 319 words without semantic content related to our analysis (e.g. friendship, love) was also removed from the dataset. In addition, specific words that refer to one or another magazine (like their names and derived neologisms) were also removed.

For the Topic Modelling (LDA), each word was replaced with its stem form (a short version of the word with suffixes removed (Bird et al., 2009)), and after the topic modelling step, a de-stemming was performed. Considering that in the present study the frequency analysis was implemented as a control, the stemming and de-stemming procedure was not performed for this case.

## RESULTS

### 1. Automatic topics detection

To evaluate the results from Topic Modelling, we run the LDA model over the full corpus, i.e., both women-oriented and men-oriented magazines, for 100 topics. We extracted the ten most relevant words for each topic (Supplementary Table 1). After careful analysis of the list of emergent topics, we selected and labelled six of them to perform the following experiments based on the relevance of these topics on gender stereotypes (Table 1).

Table 1. Selected Topics extracted from LDA analysis, manually assigned tags and ten most probable words automatically detected in each topic. The words underlined appeared originally in English in the articles. In italic, a translation for the words originally in Spanish is provided.

| id | Assigned Tag | Top 10 words |
|----|--------------|--------------|
| 1 | Women as sex object | natalia <u>hot</u> ana emma romina <u>versus</u> diez (*ten*) camilo <u>morochas</u> (*brunettes*) mega |
| 4 | Business | <u>empresa</u> (*company*) redes (*networks*) sistema (*system*) comprar (*to buy*) productos (*products*) mercado (*markets*) traves (*crossing*) tecnologia (*technology*) permite (*it allows*) desarrollo (*development*) |
| 7 | Children | niños (*kids*) adultos (*adult*) educativo (*educative*) colegio (*school*) chiquito (*tiny*) padre (*father*) secuestro (*kidnap*) <u>change</u> <u>sauna</u> pegote (*goop*) |

| 21 | Fashion | moda (*fashion*) diseño (*design*) estilo (*style*) marca (*brand*) colección (*collection*) ropa (*cloth*) tendencia (*trend*) prendas (*garments*) rosa (*pink*) zapatillas (*sneakers*) |
|----|---------|------|
| 50 | Family | hijos (*children*) madre (*mother*) mama (*mom*) padre (*father*) bebe (*baby*) familia (*family*) papa (*dad*) embarazo (*pregnancy*) regalo (*gifts*) años (*years*) |
| | | |
| 82 | Science | estudio (*study*) problema (*problem*) trabajo (*work*) explica (*explains*) ley (*law*) medico (*medic*) social (*social*) generar (*to generate*) desarrollo (*development*) investigar (*to research*) |

## 2. Automatic content analysis

To evaluate whether the topics addressed in each magazine have a gender related content bias, we estimated the probability of each topic for each article. Then, we compared the evolution of the topic probability over time for each magazine, using GAM smoothing (Hastie, 1990) in a percentage scale (Figure 2).

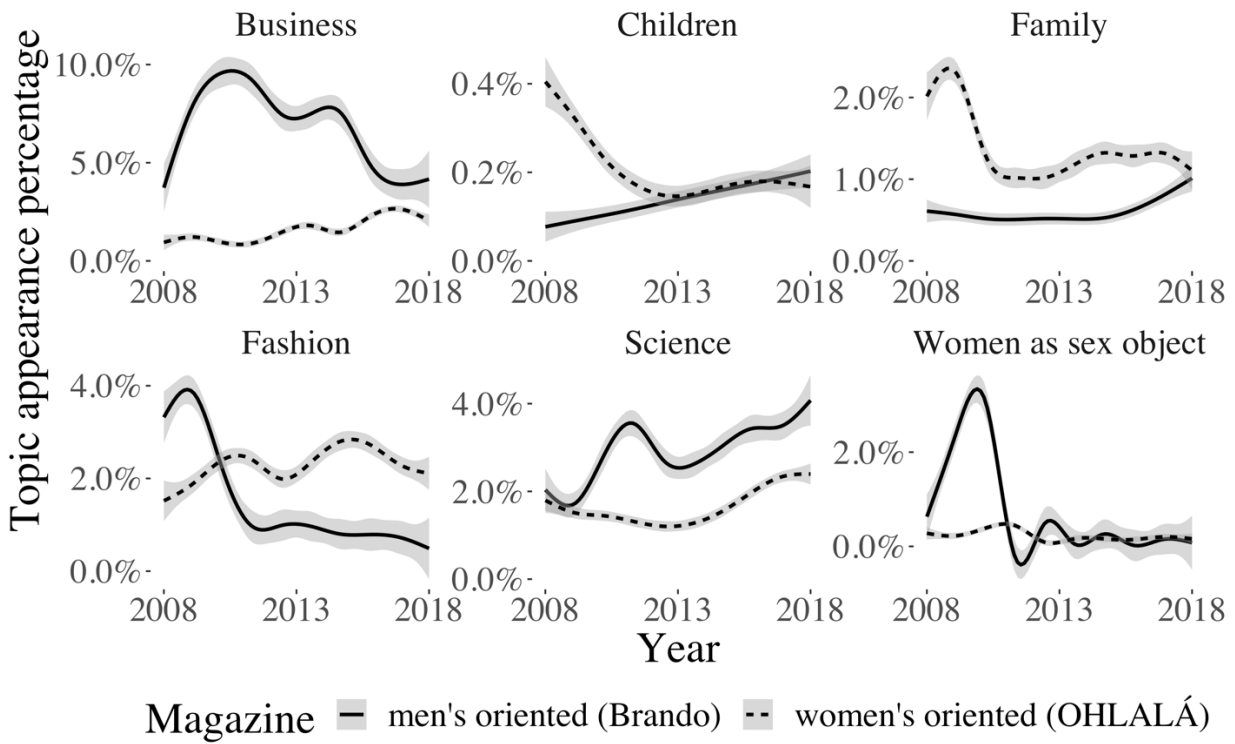

**Fig. 2**: Topic appearance percentage over time. Each colour represents a magazine. GAM smoothing of the observations (Hastie, 1990) with 95% confidence intervals in grey.

Our results show a significantly higher rate of appearance of the topic *business* in the men-oriented magazine compared to the women-oriented one, although this gap shrunk over time. The opposite occurred with the *children* topic, where at the beginning of the period analysed it presented a much stronger appearance in the women's magazine, but around 2013 this difference disappeared. The same trend was presented by the topic *family*, although the differences remained detectable until 2018. The *fashion* topic had a particular behaviour over the years. While it begins with greater representation in the men's magazine, a decreasing tendency was presented over time. In 2010 this topic was equally approached in both magazines, and then persisted to be overrepresented in the woman-oriented magazine without gap closure in 2018. One of the topics with more abrupt changes in the rate of appearance was *women as sex objects*, which had a very recurrent appearance in the early years of the men's magazine, but after peaking in 2011, an abrupt decrease is observed, and it practically disappeared. The topic *science* begins with equal presence in both magazines, however its appearance in the men-oriented magazine increased around 2010, generating a difference between the magazines that was sustained over time.

Overall, this automatic analysis displayed several differences among the content in these magazines. Most of the evaluated topics were observed to change over the years, in the women as much as in the men-oriented magazine. In addition, while some of the topics biases were reverted over time, some other topics remain associated with one of the two magazines.

### 3. Validation of automatic content analysis.

Since the topics inferred by the LDA may contain unexpected associations that come from the discursive context of the corpus (Mohr and Bogdanov, 2013), we supported these results performing a Word Frequency Analysis on the dataset.

To perform the frequency analysis, we manually selected non-ambiguous words that defined each of the six topics of interest (Table 2).

Table 2.  Words selected to represent each topic in the frequency analysis. The words underlined were used in english. In italic, a translation for the words used in spanish.

| Topic | Words |
|---|---|
| **Family** | hijos (*children*), madre (*mother*), mamá (*mom*), padre (*father*), bebé (*baby*), familia (*family*), papá (*dad*) |
| **Children** | niños (*kids*), adulto (*adult*), colegio (*school*) |
| **Business** | empresa (*company*) |
| **Fashion** | ropa (*cloth*), diseño (*design*), estilo (*style*) |
| **Science** | ciencia (*science*) |
| **Women as sex objects** | hot , morocha (*brunette*) |

For each topic we considered the ten words resulted from the LDA analysis (Table 1) and excluded the ambiguous or non-representatives. For the *science* topic we use only the word science itself, because we consider that the words within this topic might be found in other contexts not related to science.

Figure 3 shows the evolution of the *relative word frequency* compared to the total words used in each magazine each year (Equation 2). In the majority of cases, the frequency of occurrence of the most representative words of each topic reproduced the trend observed in Figure 1.

The exception to this is the frequency analysis of words associated with *children* topic, which behaviour is more similar to the *family* topic than to its own topic. This result can be interpreted as either that this topic is not coherent enough or that we erroneously label the topic. This is an example of how this verification step is important when using Topic Modelling.

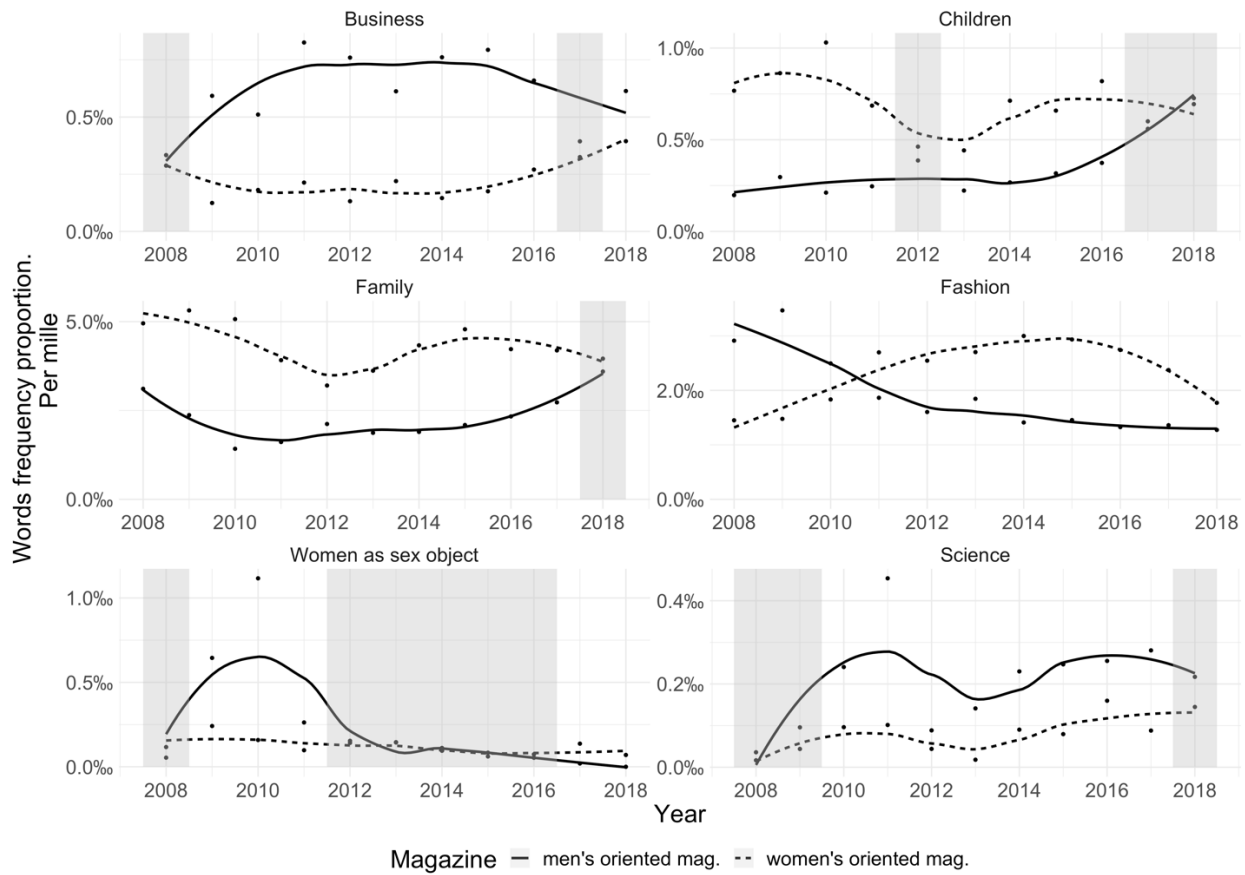

**Fig. 3**: Frequency evolution of the words in the word-lists. Solid lines are a LOESS smoothing of observations. The years with non-significant differences between the magazines are shaded in grey.

Figure 4 shows the frequency evolution of the word-list containing the word *horóscopo* (horoscope) and non-polysemic zodiac signs [horoscopo (horoscope), Tauro (Taurus), Aries, Géminis (Gemini), Escorpio (Scorpio), Sagitario (Sagittarius), Capricornio (Capricorn), Acuario (Aquarius), Piscis (Pisces) and Virgo]. We also show the frequency evolution of the word *aborto* (abortion), and the frequency evolution of words associated with feminism [feminista (feminist), feministas (feminists) and feminismo (feminism)].

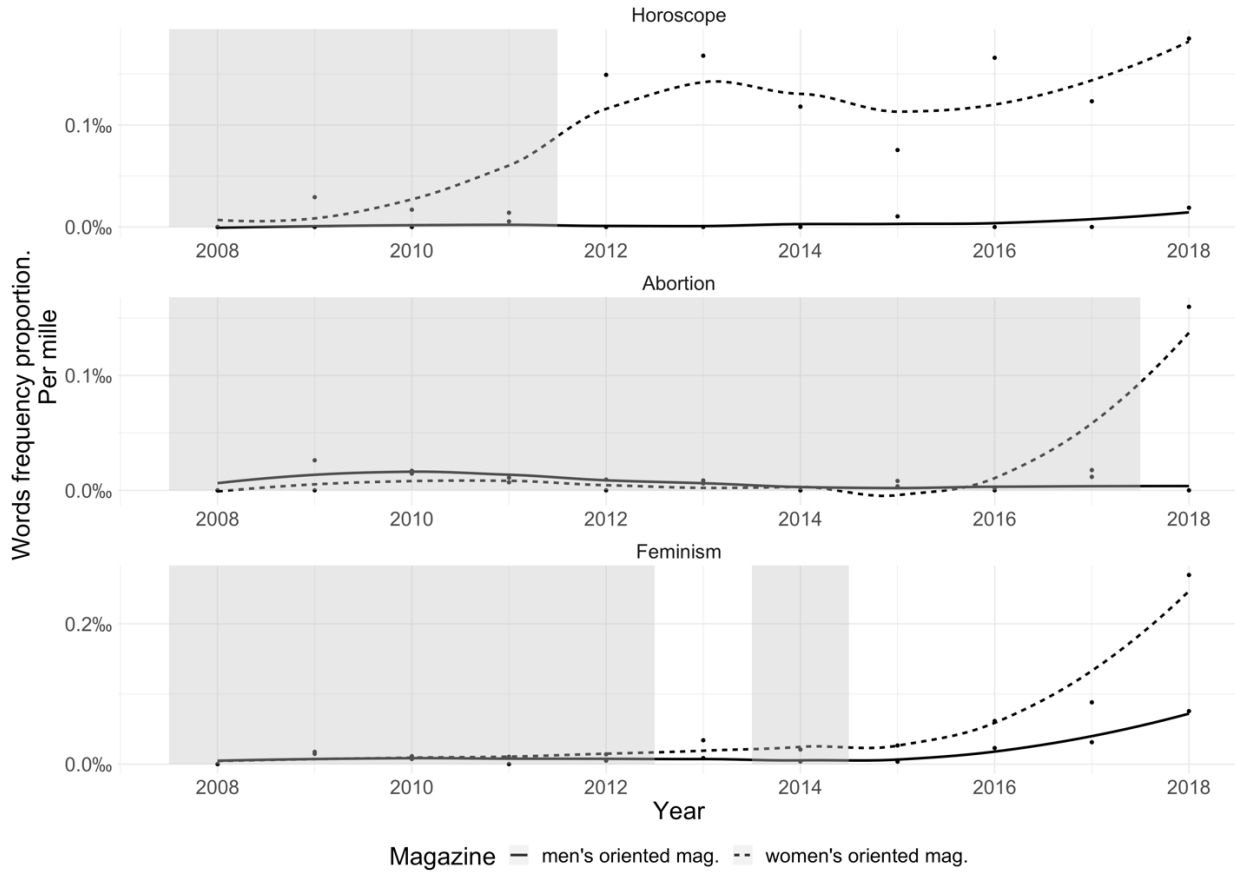

**Fig. 4**: Frequency evolution of words for **a:** Horoscope, **b:** Abortion and **c:** Feminism. Solid lines are a LOESS smoothing of observations. The years with non-significant differences between the magazines are shaded in grey.

The frequency of words associated with *horoscope* is very high and keeps increasing over the years in women-oriented magazine. On the contrary, it remains at low values in the men-oriented magazine (Figure 4a). The word *abortion* appears very infrequently in both magazines, with the exception of 2018, when the frequency of this word is significantly higher in the women-oriented magazine than in the men-oriented magazine (Figure 4b). Finally, regarding the words associated with *feminism*, an increase is observed in the appearance of them in both magazines since 2015. However, these words were more represented in the magazine oriented to women than in the one oriented to men (Figure 4c).

## DISCUSSION

The concept of gender identity is an operational term that pretends -in a political process- to enlarge visibility and legitimacy (Butler 1998). The representation has itself a normative function of a language that shows or distorts what is considered true about a category. Gender, says Butler, "is not always constituted coherently or consistently in different historical contexts, because gender intersects with racial, class, ethnic, sexual and regional modalities of discursively constituted identities" (Butler 1998).

Several studies have analysed differential content in gender-oriented magazines. One of the first texts that made a comparison of female and male press is Courtney and Lockeretz (1971), that analysed seven magazines that were directed toward both male and female readers. Over the years, these analyses were performed mainly by manual content quantification, presenting a limitation on the number of articles that were included.

Using a Topic Modelling technique, we were able to successfully analyse 24,000 articles from two magazines, one oriented to men and the other oriented to women, in a period of 10 years. By complementing this model with Word Frequency Analysis, we were able to automatically identify five relevant topics (*business, family, women as sex objects, fashion* and *science*). This method allows discovering themes that are present in the corpus and represents each theme according to the probability of occurrence of each word in it. Unlike the classic method based on static word lists, the words that represent the topics are specific to the historical moment and the environment, thus solving the concerns raised by Zotos and Tsichla (2014) about the effects of strong cultural changes.

Our analysis shows that there are differences in the frequency of appearance of topics between both journals, and that these differences were consistent with well-studied gender stereotypes that reinforce traditional social roles (Murillo et al, 2010).

Our findings are aligned with the literature regarding the oversexualization of women (Downs and Smith, 2010), portrait of women as responsible for family matters (Das 2011; Coltrane, 1997), association of science and business with men and not to women (Matud et al., 2011; Courtney and Lockeretz, 1971) and association of fashion with women (Murillo et al, 2010, Davalos 2007, Baile 2020). Despite the fact that there are studies that characterize the content of horoscopes in magazines oriented to women and girls (Tandoc and Ferrucci 2014), little research has been carried out to address the difference between media oriented to men and women, regarding of the content associated with horoscope and astrology.

Our results showed that there was a strong tendency to close the content distribution gap between the magazines on topics such as *business*, *family*, and *women as sex objects.* The latter topic fell abruptly between 2011 and 2012. The timing of this change has a strong coincidence with the irruption of the feminist movement in political arena of Argentina, starting with the first "Marcha de las putas" (the Slutwalk) in 2011 (Franchini Díaz y Pastor, 2012). Regarding the *family* topic, our finding is aligned with the signs observed by Marshall et al (2014) of a change in the representation of fatherhood in magazine advertisements. On the other hand, the *fashion* and *science* gap shows no signs of narrowing. In addition, the gap in the content associated with the *horoscope* shows an increase when compared to the beginning of the analysed period.

Since 2015 there has been a sharp increase in the presence of feminism on the magazine's pages aimed at women. This effect coexists with the rise of the #NiUnaMenos movement and its presence in the media. This movement protests against gender violence and femicides (O'Dwyer 2016; Davies, 2017; Garibotti and Hopp 2019). In 2018, a jump is observed in the use of the word *abortion* in the magazine aimed at women, reflecting the uprising debate on the legalization of abortion in Argentina that was discussed in Congress during that year (Lucaccini et al 2019).

Also, performing a word frequency analysis as a control turned out to be good practice. This was evidenced in the Word Frequency Analysis of the topic *children,* which did not show a behaviour similar to that of Topic Modelling, so we can disregard it from the analysis.

Overall, our work presents a novel application of methods that proves to be useful in analysing large amounts of articles with computational tools and that contribute to the field of gender bias analysis in magazine content.

## 1. Limitations

The methodological framework proposed in this study focuses on the textual evidence of gender stereotypes in magazines. A shortcoming of this approach is that it doesn't consider the non-textual evidence, such as images and other formats. The visual content on magazines plays a vital role in gender stereotypes reproduction, especially in advertising (Kahlenberg and Hein 2010, Mager and Helgeson 2011, Nam et al. 2011). An approach that considers both the textual and non-textual dimensions could be in this sense enriching.

Regarding Topic Modelling, it automatically captures the most frequent topics, and it is worth considering that some of these might be difficult to contextualize and comprehend. Therefore, results interpretation and topic curation by an expert from the field is required, making the method not fully automatic. Furthermore, relevant issues or the hypothesis to be tested may not be well represented in the automatically generated topics (e.g. horoscope, abortion and feminism), thus, the analysis of word frequency can be a good complement.

On the word frequency approach, it is important to note that conducting many tests might lead to misleading conclusions due to the problem of multiple comparisons. Therefore, the lists need to be carefully constructed and theoretically justified before experimentation, highlighting once again the relevance of interdisciplinary work among computer and social scientists in the experimental design and interpretation.

## 2. Future research directions

In the present research, our source of data were two magazines from the same editorial group. In order to have a complete understanding of the gender roles in magazines, it would be interesting to extend the data sources to a more general corpus. Also, the analysis is focused on Argentinean magazines. Replicating these methods on magazines from different countries would allow international comparison of gender stereotypes. Also, our study covers the years 2008 to 2018. In future work, we expect to expand this range to the present.

### 3. Practical implications

During the last few years, a great improvement in Argentina in terms of awareness of gender biases, both in society and in the media, has been made. This is strongly represented in the results of our work for some of the analysed topics, such as *women as sex objects, family,* and *business.* However, certain differences between the magazines oriented to men or women persist and have not shown evolution during the analysed period - on topics such as *fashion, science,* and *horoscope.* Therefore, our results show not only the topics that are currently under discussion in Argentine society, but also might enlighten those that have not yet emerged in public debate. Besides, both the proposed methodology and the results themselves contribute to the open discussion. Furthermore, as the methodological framework proposes an automated way to track the evolution of gender stereotypes in magazines, it might be an interesting and useful tool for the local feminist movement to provide quantitative evidence of this problem.

### CONCLUSION

The goal of this study was to use Data Science and Natural Language Processing techniques to compare the content of a women-oriented magazine with that of a men-oriented produced by the same editorial over a decade (2008-2018). Both magazines contained in total more than 24,000 articles, and, although there are several studies that use traditional methods to quantify the content of women and men-oriented magazines, as far as we know, ours is the first study that compares magazines in such a large dataset.

Our approach was to use Topic Modelling to identify the main topics discussed in the magazines, and then, quantify how much the presence of these topics differs between magazines over time. This method allowed us to discover content present in the articles and represent the topics according to the words that are most associated with them. Also, a validation step was carried out by performing a Word Frequency Analysis. This test validated five of the six topics: *business*, *family*, *women as sex objects*, *fashion* and *science.*

Overall, our work contributes to the longstanding field of gender content analysis in mass media (Kuipers et al., 2017; Collins, 2011; Rudy, 2010 Murillo et al, 2010; Eisend, 2010; Gill, 2009; Kress & van Leeuwen, 2006; Koivula, 1999) but with a novel methodology that highlights the relevance of interdisciplinary work and shows how computational tools are useful resources to analyse large amounts of content among platforms.

**References**

Baile, J. I., Gabino-Camposa, M., & Pérez-Lugoa, A. L. (2020). Analysis of the aesthetic stereotypes of women in nine fashion and beauty Mexican magazines. *Revista Mexicana de Trastornos Alimentarios, 7*(1), 40-45.

Bandura, A., Ross, D., & Ross, S. A. (1963). Imitation of film-mediated aggressive models. The Journal of Abnormal and Social Psychology, *66*(1), 3–11.

Bird, S., Klein, E., & Loper, E. (2009). *Natural language processing with Python: analyzing text with the natural language toolkit.* O'Reilly Media, Inc.

Blei, D. M., Ng, A. Y., & Jordan, M. I. (2003). Latent dirichlet allocation. *Journal of machine Learning research, 3*(Jan), 993-1022.

Butler, J., & Lourties, M. (1998). Actos performativos y constitución del género: un ensayo sobre fenomenología y teoría feminista. *Debate feminista, 18*, 296-314.

Chang, J., Gerrish, S., Wang, C., Boyd-Graber, J. L., & Blei, D. M. (2009). Reading tea leaves: How humans interpret topic models. In *Advances in neural information processing systems, 22*, 288-296.

Collins, R. L. (2011). Content analysis of gender roles in media: Where are we now and where should we go?. *Sex roles, 64*(3-4), 290-298. doi:10.1007/s11199-010-9929-5.

Coltrane, S., & Adams, M. (1997). Work-family imagery and gender stereotypes: Television and the reproduction of difference. *Journal of Vocational Behavior, 50*, 323–347.

*Comercial LA NACION*. [online] Available at:

<http://comercial.lanacion.com.ar/pages/productocomercial.aspx?cat=revista>
[Accessed 23 November 2020].

Courtney, A. E., & Lockeretz, S. W. (1971). A woman's place: An analysis of the roles portrayed by women in magazine advertisements. *Journal of Marketing Research*, *8*(1), 92-95.

Das, M. (2011). Gender role portrayals in Indian television ads. *Sex Roles*, *64*(3-4), 208-222. doi:10.1007/s11199-010-9750-1.

Davalos, D. B., Davalos, R. A., & Layton, H. S. (2007). III. Content analysis of magazine headlines: Changes over three decades?. *Feminism & Psychology*, *17*(2), 250-258.doi:10.1177/0959353507076559.

Davies, M. C. (2017). Nos Están Matando [They are Killing Us]: Feminist Movements' Influence in Argentina and Chile. Zenith. *Undergraduate Research Journal for the Humanities at the University of Kansas*, *2*(1), 165-182.

Del-Teso-Craviotto, M. (2006). Words that matter: Lexical choice and gender ideologies in women's magazines. *Journal of Pragmatics*, *38*(11), 2003-2021., doi:10.1016/j.pragma.2005.03.012.

Desmond, R., & Danilewicz, A. (2010). Women are on, but not in, the news: Gender roles in local television news. *Sex Roles*, *62*(11), 822-829., doi:10.1007/s11199-009-9686-5.

Downs, E., & Smith, S. L. (2010). Keeping abreast of hypersexuality: A video game character content analysis. *Sex roles*, *62*(11-12), 721-733. doi:10.1007/s11199-

009-9637-1.

Díaz, G. F., & Pastor, C. (2012, December). Tu mamá también: Apropiaciones de la Marcha de las Putas en Argentina. In *Actas del 2° Congreso Interdisciplinario sobre Género y Sociedad:"Lo personal es político"* (Vol. 1, No. 1).

Eisend, M. (2010). A meta-analysis of gender roles in advertising. *Journal of the Academy of Marketing Science*, *38*(4), 418-440.

Foucault, M. (1978). *The history of sexuality: volume I-An introduction.* Pantheon Books.

Gálvez, R. H., Tiffenberg, V., & Altszyler, E. (2019). Half a Century of Stereotyping Associations Between Gender and Intellectual Ability in Films. *Sex Roles, 81(9-10), 643-654.*

Garibotti M.C., Hopp C.M. (2019) Substitution Activism: The Impact of #MeToo in Argentina. In: Fileborn B., Loney-Howes R. (eds) #MeToo and the Politics of Social Change. Palgrave Macmillan, Cham.

Gill, R. (2009). Beyond the 'sexualization of culture' thesis: An intersectional analysis of 'sixpacks', 'midriffs' and 'hot lesbians' in advertising. *Sexualities, 12*(2), 137–160. https://doi.org/10.1177/1363460708100916

Gilpatric, K. (2010). Violent female action characters in contemporary American cinema. *Sex Roles*, *62*(11-12), 734-746. doi:10.1007/s11199-010-9757-7.

Goffman, E. (1979). *Gender advertisements.* Macmillan International Higher Education.

Hastie, T. J., & Tibshirani, R. J. (1990). *Generalized additive models* (Vol. 43). CRC press.

Hether, H. J., & Murphy, S. T. (2010). Sex roles in health storylines on prime time

television: A content analysis. *Sex Roles*, *62*(11-12), 810-821. doi:10.1007/s11199-009-9654-0.

Kahlenberg, S. G., & Hein, M. M. (2010). Progression on Nickelodeon? Gender-role stereotypes in toy commercials. *Sex Roles*, *62*(11-12), 830-847. doi:10.1007/s11199-009-9653-1.

Koivula, N. (1999). Gender stereotyping in televised media sport coverage. Sex roles, *41*(7-8), 589-604.

Kress, G., & van Leeuwen, T.(2006) *Reading images: The grammar of visual design.* Second edition, London: Routledge.

Kuipers, G., Van Der Laan, E., & Arfini, E. A. (2017). Gender models: changing representations and intersecting roles in Dutch and Italian fashion magazines, 1982–2011. *Journal of Gender Studies, 26*(6), 632-648.

Lama, M. (2013). *El género. La construcción cultural de la diferencia sexual.* México: Librero.

Lucaccini, M., Zaidán, L., & Pecheny, M. (2019). Qué nos dice el debate sobre aborto en 2018 sobre la clase política y el espacio público en la Argentina. *A. Moreno, D. Maffía y PL Gómez (Comps.), Miradas feministas sobre el derecho. Buenos Aires: Editorial jusbaires*, 245-263.

Magazine links (accessed in 2018) brando url: https://www.lanacion.com.ar/revista-brando; Ohlala url: https://www.lanacion.com.ar/revista-ohlala

Mager, J., & Helgeson, J. G. (2011). Fifty years of advertising images: Some changing perspectives on role portrayals along with enduring consistencies. *Sex Roles, 64*(3-

4), 238-252. doi:10.1007/s11199-010-9782-6.

Murillo, M. F. M., Vizuete, J. I. A., & Learreta, M. G. (2010). Claves de la construcción de género en las revistas femeninas y masculinas: análisis cuantitativo. *Estudios sobre el Mensaje Periodístico*, *16*, 259-289.

Marshall, D., Davis, T., Hogg, M. K., Schneider, T., & Petersen, A. (2014). From overt provider to invisible presence: discursive shifts in advertising portrayals of the father in Good Housekeeping, 1950–2010. *Journal of Marketing Management*, *30*(15-16), 1654-1679.

Matud, M. P., Rodríguez, C., & Espinosa, I. (2011). Gender in Spanish daily newspapers. *Sex Roles*, *64*(3-4), 253-264. doi:10.1007/s11199-010-9874-3.

Michel, J. B., Shen, Y. K., Aiden, A. P., Veres, A., Gray, M. K., Pickett, J. P., ... & Pinker, S. (2011). Quantitative analysis of culture using millions of digitized books. science, *331*(6014), 176-182.

Mohr, J. W., & Bogdanov, P. (2013). Introduction—Topic models: What they are and why they matter. *Poetics*, *31*, 545-569.

Nam, K., Lee, G., & Hwang, J. S. (2011). Gender stereotypes depicted by Western and Korean advertising models in Korean adolescent girls' magazines. *Sex Roles*, *64*(3-4), 223-237. doi:10.1007/s11199-010-9878-z.

Neuendorf, K. A. (2011). Content analysis—A methodological primer for gender research. *Sex Roles*, *64*(3-4), 276-289.

Neuendorf, K. A., Gore, T. D., Dalessandro, A., Janstova, P., & Snyder-Suhy, S. (2010). Shaken and stirred: A content analysis of women's portrayals in James Bond films.

*Sex Roles, 62*(11-12), 747-761. doi:10.1007/s11199-009-9644-2.

O´Dwyer, M. (2016). #NiUnaMenos; standing up to femicides and "machismo" in Argentina. Retrieved october 2020, from https://genderandpoliticsucd.wordpress.com/2016/02/03/niunamenos-standing-up-to-femicides-and-machismo-in-argentina/.

Rudy, R. M., Popova, L., & Linz, D. G. (2010). The Context of Current Content Analysis of Gender Roles: An Introduction to a Special Issue. *Sex roles, 62*(11-12), 705-720.

Scott, J. W. (1986). El género: una categoría útil para el análisis histórico'. *Historical review, 91*, 1053-1075.

Serret, E. (2001). *El género y lo simbólico: la constitución imaginaria de la identidad femenina*. Universidad Autónoma Metropolitana-Azcapotzalco.

Spataro, C. (2013). Las tontas culturales: consumo musical y paradojas del feminismo. *Revista Punto Género*, (3), ág-27. doi:10.5354/0719-0417.2013.30265.

Tandoc Jr, E. C., & Ferrucci, P. (2014, July). So says the stars: A textual analysis of Glamour, Essence and Teen Vogue horoscopes. In *Women's Studies International Forum* (Vol. 45, pp. 34-41). Pergamon. doi:10.1016/j.wsif.2014.05.001.

Turner, J. S. (2011). Sex and the spectacle of music videos: An examination of the portrayal of race and sexuality in music videos. *Sex Roles, 64*(3-4), 173-191. doi:10.1007/s11199-010-9766-6.

Zotos, Y. C., & Tsichla, E. (2014). Female stereotypes in print advertising: A retrospective analysis. *Procedia-social and behavioral sciences, 148*, 446-454.